\documentclass{article}

\usepackage[english]{babel}
\usepackage{authblk}
\usepackage[letterpaper,top=2cm,bottom=2cm,left=3cm,right=3cm,marginparwidth=1.75cm]{geometry}
\usepackage{amsmath}
\usepackage{graphicx}
\usepackage[colorlinks=true, allcolors=blue]{hyperref}
\usepackage{amsfonts}
\usepackage{multirow}
\usepackage{xr}
\newcommand{\cT}{\mathcal{T}}
\newcommand{\cB}{\mathcal{B}}
\newcommand{\cM}{\mathcal{M}}
\newcommand{\cC}{\mathcal{C}}
\newcommand{\cD}{\mathcal{D}}
\newcommand{\cU}{\mathcal{U}}
\newcommand{\cA}{\mathcal{A}}
\newcommand{\bx}{\mathbf{x}}
\newcommand{\bdelta}{\mathbf{\delta}}

\newcommand{\br}{\mathbf{r}}
\newcommand{\bs}{\mathbf{s}}
\newcommand{\cCb}{\mathbf{\cC}}
\newcommand{\R}{\mathbb{R}}

\newcommand{\Var}{\text{Var}}
\newcommand{\at}{\widetilde{a}}

\title{From slides (through tiles) to pixels: an explainability framework for weakly supervised models in pre-clinical pathology}
\author[1]{Marco Bertolini \thanks{Corresponding author: marco.bertolini@bayer.com}}
\author[2]{Van-Khoa Le \thanks{Current Affiliation:
Median Technologies, 1800 Rte des Crêtes Batiment B, 06560 Valbonne, France}}
\author[1]{Jake Pencharz}
\author[1]{Andreas Poehlmann}
\author[1]{Djork-Arné Clevert \thanks{Current Affiliation:
Machine Learning Research, Pfizer Worldwide Research Development and Medical, Linkstr.10, Berlin, Germany}}
\author[1]{Santiago Villalba}
\author[1]{Floriane Montanari}
\affil[1]{Machine Learning Research, Data Science\&AI, Bayer AG, Berlin, Germany}
\affil[2]{Bayer Crop Science, 355 Rue Dostoievski 06560 Valbonne}

\begin{document}
\maketitle

\begin{abstract}
In pre-clinical pathology, there is a paradox between the abundance of raw data (whole slide images from many organs of many individual animals) and the lack of pixel-level slide annotations done by pathologists. Due to time constraints and requirements from regulatory authorities, diagnoses are instead stored as slide labels. Weakly supervised training is designed to take advantage of those data, and the trained models can be used by pathologists to rank slides by their probability of containing a given lesion of interest. 

In this work, we propose a novel contextualized eXplainable AI (XAI) framework and its application to
deep learning models trained on Whole Slide Images (WSIs) in Digital Pathology. Specifically, we apply our methods to a multi-instance-learning (MIL) model, which is trained solely on slide-level labels, without the need for pixel-level annotations. We validate quantitatively our methods by quantifying the agreements of our explanations' heatmaps with pathologists' annotations, as well as with predictions from a segmentation model trained on such annotations.
We demonstrate the stability of the explanations with respect to input shifts, and the fidelity with respect to increased model performance. We quantitatively evaluate the correlation between available pixel-wise annotations and explainability heatmaps.
We show that the explanations on important tiles of the whole slide correlate with tissue changes between healthy regions and lesions, but do not exactly behave like a human annotator. This result is coherent with the model training strategy.

\end{abstract}

\section{Introduction}

In pharmaceutical research, compounds progressing towards clinical trials must undergo stringent tests for their safety and efficacy before the first human exposure. This is done in a pre-clinical setting, where laboratory animals (very often rodents) are exposed to the new drug candidate under various dosing regimen. Such pre-clinical trials are highly regulated by the regulatory agencies, and must follow Good Laboratory Practices (GPL). The Food and Drugs Administration (FDA) publishes guidance for industry on appropriate toxicology studies \cite{fda-2008}. The well-known repeated dose toxicity studies, for example, must take place in two mammal species (one non-rodent), for a duration that should be equal to or exceed the planned human clinical trial length (up to nine months in the non-rodent species). Animals of both genders are divided in different groups, with some (receiving no treatment) taking the role of control groups. At the end of the study, all animals are sacrificed, their organs weighted and tissue samples of all relevant organs are collected for histopathological analysis \cite{HAYES2020}. Such a study may involve over a hundred animals, a dozen organ types, with many samples taken for each organ. Collected tissues are trimmed following organ-specific guidance \cite{Organ-trimming-2003}, then embedded in paraffin, sectioned and mounted on glass slides. For histopathology visualization, slides are then typically stained with hematoxylin and eosin (H\&E), and evaluated one by one under microscope by certified pathologists \cite{pitchford2018}.

The recent advances of machine learning in computer vision, together 
with faster whole slide image (WSI) scanning procedures, 
promise to alleviate the most significant challenges in the field of pathology: an overall decreasing pathologist workforce coupled with an increasing demand for diagnostic throughput \cite{KOMURA201834}.
However, the successful implementation of a large-scale digital pathology workflow comes with additional hurdles \cite{ai_cp_cui}. 
The datasets are often remarkably large (one WSI has a size of $\sim$1-2GB), requiring adequate storage and logistics solutions, and are characterized by inhomogeneous and multi-modal metadata and ground-truth (e.g., slide-level diagnosis, pixel-level annotations).
Machine learning models can support pathologists in several tasks, such as detecting 
the presence of lesions on H\&E-stained slides, classification of lesion grading, pixel-wise segmentation of lesion areas, and immunohistochemistry quantification \cite{Iizuka2020, Yan2020, Chen2016, Sheikzadeh2018}. 
The feasibility of these tasks differ according to the amount of data available for training and their level of curation, for instance, whether pixel-wise segmentations are available. 
In the standard evaluation workflow, a pathologist would look at the slides under the microscope (or using a purpose-built visualization software if the slides were digitized) and deduce one or several diagnoses (e.g. "this slide contains signs of hyperplasia"). Such a slide-level diagnosis is fast to obtain if the pathologist is a trained expert. This is the information that is historically stored as metadata together with all the slides for all pre-clinical studies.
In some ad-hoc cases, pathologists may take the effort to actually delineate the regions where the lesions are found, using digital slide annotation tools. This is time-consuming, error-prone, and not all lesions will be annotated systematically. This is what we refer to as pixel-wise segmentations.

Within a typical pre-clinical study, most samples will not show any lesion. Therefore, a lot of the diagnosis time is spent on non-problematic slides. A digital application that would be able to prioritize slides for review, proposing first the samples most likely to contain lesions, would help pathologists spend more of their time on interesting slides. Furthermore, it would be desirable for such a system to also highlight the most likely sub-regions where a particular lesion occurs. The combination of a powerful whole slide ranking system with patch-level highlights could be obtained in two ways.

First, methods based on fully-supervised feature localization intrinsically satisfy this requirement (i.e., part of their output is a segmentation mask on top of the slide) but require a large number of pixel-level manual annotations to train on, which are very costly to obtain. 
On the other hand, weakly-supervised models only require slide-level diagnosis and, provided that enough data is available, achieve strong performance in classification and detection tasks \cite{campanellaMIL}. Such learning strategies do not require additional work from highly skilled pathologists since the model is trained with (weak) labels already generated in the regular study-reading process. However, such benefit comes at the cost of reduced interpretability of the model's predictions, since the default output is a score for the whole slide without fine-grained details.

The large size of the WSIs is such that it is not possible to feed whole scans
to the machine learning algorithm, at least not at the necessary resolution
for detection of the meaningful morphological features. 
The most common approach is to subdivide the slide into smaller fixed-sized tiles (or patches) according to a rectangular grid, and feed only these smaller images (typically on the order of a few hundred pixels
per side) to the algorithm.
This step does not introduce difficulties for fully manually annotated slides, since a pixel-level ground truth is automatically inherited by each tile.
If only the slide-level diagnosis are known, 
a positive diagnosis is assigned to the slide if and only if the lesion is present somewhere in the full image. That is, the lesion is present in at least one of its tiles, while others, often the majority, will only exhibit healthy tissue. 
This formulation, with a bag of elements (here, the tiles) for which only the bag label is known (here, the whole slide diagnosis), is known as the Multi-Instance problem \cite{Dietterich1997SolvingTM, Andrews2002MultipleIL}. 
Machine Learning models targeted at learning under this paradigm 
are known as Multi Instance Learning (MIL) models, and have
been adopted in various areas of machine learning,
with particular success in computer vision applications \cite{Viola2005MultipleIB, Zhang2001EMDDAI, Kraus2016ClassifyingAS}.
The typical architecture consists in a classifier, whose goal is to solve the task on individual bag elements, and a pooling strategy which predicts the global bag label given the collection of individual element predictions.

In this work, we focus on rat liver necrosis. We build an explainability framework for weakly-supervised models that localizes the predictions to pixel-level annotation-like heat maps. We apply our methods to a MIL model trained on both public and internal data with slide-level diagnosis. 
We argue that the heatmaps we generate give a strong guidance into understanding how the model learned to solve the task. 
We validate quantitatively our explanations by computing correlation between our explainability heatmaps and ground truth manual annotations which were not used for model building. We also suggest to utilize a segmentation model trained on such annotations to generate a larger benchmark set
for the explanations. This can be particularly useful, for example, to monitor model performance and model 
interpretability over time, without the need for additionally generating new manual annotations. 
Finally, we show that our explanations are robust with respect to the tiling grid choice.
That is, we show that the explanations from separate but overlapping tiles tend to agree
on the overlap area. 
Taken together, our results show that we can make use of commonly obtained slide-level diagnosis to build a digital pathology assistant that can rank slides by likelihood of containing a necrosis lesion, and, using explainability techniques, further inform the pathologist on the areas of interest and relevant morphological features according to the model.
 We show that, rather than fully overlapping with annotations, our explanations seem to focus on the boundaries between healthy and pathological tissue. This is probably the strategy employed by the MIL model to detect tiles containing necrosis lesions.

\subsection{{Related Work}}
Weakly supervised learning is a commonly used paradigm in digital pathology, due to the combination of very large images and lack of pixel-resolution labels. Multiple instance learning was used by Hou and colleagues \cite{Hou2016} to recognize cancer subtypes in the TCGA dataset. Campanella and colleagues \cite{campanellaMIL} also compare the use of MIL with building a fully supervised model on a small annotated subset, and show that, as could be expected, weak supervision on large datasets leads to better generalization capability.  

In a recent work, Hägele and colleagues \cite{haegele2020} explore how explainability may help relieve some challenges in digital pathology. The challenges are, namely, noisy labels, staining variability, imbalanced data and lack of labels. They build a fully supervised tile classifier (i.e., they have tile-level labels), and apply an explainability algorithm, layer-wise relevance propagation \cite{LRP}, to obtain saliency maps at the tile level. Such maps are then qualitatively and quantitatively compared with manual annotations by pathologists. Note that this work does not make use of weakly supervised models.

\begin{table}[t!]
    \centering
    \begin{tabular}{c| c c c p{0.13\textwidth}}
         Model & Dataset & \# WSIs &  \# Tiles & 
         \# Necrosis Annotations
         \\ \hline
         \multirow{ 2}{*}{Segmentation} 
         &Train  & 185  & 17k 
         & 2004 \\
         &Test   & 45   & 6k  
         & 2048\\
         \hline
        \multirow{ 2}{*}{MIL} 
         &Train  & 3519 &-  & - \\
         &Test   & 464 & - & - \\
         \hline
    \end{tabular}
    \caption{Left: Subset of TGGates and Bayer data used to train/validate the segmentation and MIL models.}
    \label{tab:segmentation_data}
\end{table}

Lu and colleagues \cite{CLAM} propose a modification to the standard MIL concept, the clustering-constrained attention multiple instance learning (CLAM). In their approach, an attention mechanism is used to pool signal from the most interesting tiles. They use those attention scores on the tile level as a way to get explainability on the slide level: important patches according to the model will be highlighted by high attention scores. Such coarse-grained heatmaps can be further refined by sliding the tiling with overlap. This of course is a computationally expensive addition, where the same tissue must be sent over many times over for inference. The authors recommend CLAM for cases where the training data is limited, showing great prediction performances with merely a few hundred whole slide images. 
Huang and Chung \cite{CELnet} combined two XAI methods,  class activation maps (CAM) and saliency to input pixels to obtain explanations at different resolution levels for a weakly supervised model with attention. They test whether the obtained saliency maps co-localize with annotated cancer areas in their dataset. While conceptually similar to our approach, this study designs a very specific XAI workflow that is tailored to the model architecture.
In this work, we propose a general explainability framework that could work with many implementations of the MIL concept.
In fact, our methods apply potentially to any learning task whose network's architecture
is a Convolutional Neural Network (CNN). The key feature of our framework is that it includes information from different resolution levels. 

\section{Methods}

\subsection{Dataset}

Our dataset consists of $\sim$4500 WSIs from the Open TG-GATEs dataset \cite{igarashi2015open}, as well as from Bayer's internal studies. In total, we collected slides from 228 TG-GATEs studies and 13 Bayer's internal studies.
For this work, we selected liver sections of rats, and we focus on the diagnosis \emph{necrosis}.
Liver is the primary organ responsible for detoxification of pharmaceuticals compounds, 
while rats have been the most-adopted rodent in pre-clinical toxicological assessments.
All the slides were stained with H\&E (hematoxylin and eosin).
Of all the slides, about 20\% are assigned the diagnosis necrosis.
The splits are performed study-wise: the different sets contain slides originating
from different pre-clinical studies using different drug candidates. This setting is meant to increase the difficulty of the generalization
task and to mimic a practical application of the model, which is applied whenever a new
study is scanned and digitized. 
We performed 5-fold cross-validation, where for each fold we split the data in train/validation/test sets according to the ratio 80/10/10. The validation set was only used during training to 
track the generalization accuracy of the model and later for model selection. 

To train a segmentation model, in addition to images, pixel-level annotations are necessary. 
These were generated by Bayer's pathologists in a collaboration project with the 
Berlin-based company AIgnostics. The annotated dataset is available on Zenodo \cite{dataset}. 
The model is trained on a small subset of TG-GATEs images for which such annotations were obtained. 
Details of the training and testing datasets can be seen in Table \ref{tab:segmentation_data}. During training, we exclude slides without annotations, and remove any tiles which did not feature mitotic figures, necrotic cells, or areas of necrotic tissue. Furthermore, it is assumed that any pixels which were not annotated belong to the parenchyma (i.e. healthy tissue) class. This assumption helped in practice with model performance. In fact, models trained on tiles which were masked to only include annotated pixels performed substantially worse than those trained on tiles where some tissue was incorrectly assumed to be normal (data not shown). 

All our explainability results are computed on the intersection between the test sets of the
MIL and segmentation models.

\paragraph{Preprocessing steps:}

\begin{figure}[t!]
\center
\includegraphics[width=\textwidth]{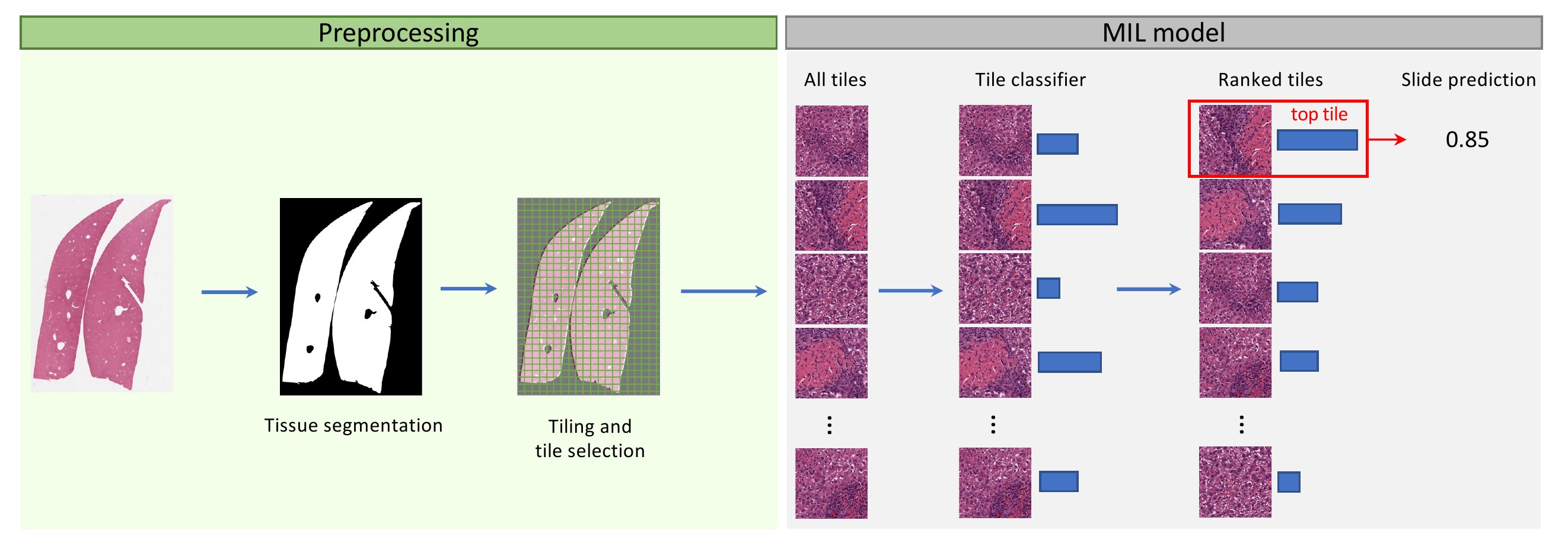}
\caption{Overview of slide preprocessing steps and inference strategy for the MIL model.}
\label{fig:MIL_overview}
\end{figure}

Figure \ref{fig:MIL_overview} depicts the general preprocessing steps. We perform tissue filtering, removing (white) background and artifacts of the slide (pen marks, etc.) using the 
library \texttt{PaDO} \cite{pado}. Then, we extract fixed-sized images by applying a grid for tiling.
We choose to apply a non-overlapping grid with tile size of $L=512$. Tiles with less than 
$80\%$ of pixels covering tissue are excluded. Tiles are extracted at the native MPP (micron per pixel) resolution.

In an effort to reduce data drift between slides collected from different labs and at different times, Macenko's stain normalisation flow \cite{macenko2009method} was used to normalise all tiles passed to the segmentation model. This normalization is now implemented in our in-house framework for 
model development \cite{palo}.

\subsection{MIL Model}

We trained a MIL model, which we denote $\Omega$, to predict whether a WSI exhibits presence or absence of necrotic area. As mentioned above, the model's strategy 
consists of a tile classifier $\Phi$ and a pooling function. Specifically, each WSI is 
subdivided into a collection of non-overlapping tiles
$\cB_A = \{T^{(A)}_1, T^{(A)}_2, \dots, T^{(A)}_{N_A}\}$,
where $\cT_i\in \cM = [0,255]^{L \times L \times 3}$. The 3 input channels correspond to the RGB color values.

The index $A$ identifies the WSI and $N_A$ denotes the slide-dependent number of tiles in the
slide. 
Each tile in a bag is passed through the classifier network $\Phi$, which assigns to it a scalar value with value in $[0,1]$, i.e., 
$\Phi(\cT^{(A)}_i) = s_i^{(A)} \in [0,1]$ for $i=1,\dots, N_A$.
Finally, we apply a max-pooling function, that is, we assign to the WSI the highest score in the tiles bag as follows
\begin{align}
\label{eq:MILstrategy}
    \Omega(\cB_A) = \max\{\Phi(\cT_1^{(A)}), \Phi(\cT_2^{(A)}), \dots, 
    \Phi(\cT_{N_A}^{(A)}))\} = \max_i\{s_i\}_{i=1,\dots,N_A}. 
\end{align}
The implemented MIL model strategy is depicted in Figure \ref{fig:MIL_overview}. 
During training, \eqref{eq:MILstrategy} implies that only the top tiles of each slide
play an active role in the back-propagation step. Which
specific tile for a given WSI gets selected as the top tile
evolves during the training as the classifier improves its performance.

In our experiments, we chose $\Phi$ to be a pretrained ResNet50, where we removed the last classification layer, 
concatenated with a 
downstream network consisting of a stack of fully connected layers with ReLU non-linearity. 
We refer to the Supplementary Information (SI) for further details concerning hyperparameters, training details and further architecture specifications. 
We train the downstream network with the MIL paradigm for a fixed amount of epochs (50), and we track generalization performance on the validation set. We use the performance on the validation set for
model and architecture selection.
Then, we evaluate the performance on the test set, using ROC AUC as our evaluation metric. 
The model exhibits an average test performance of $0.85 \pm 0.03$ (over 5 folds). In table
\ref{tab:segmodel_results} we report also precision and recall, where we take the threshold $0.5$ for
the classification task. Because we work in a toxicology setting, and the model should be used to prioritize slides for review, we want the model to be very good at detecting non-relevant slides, i.e. slides that do not contain any lesion. Therefore, in the table, precision and recall refer to the lesion-free class: in our test set, 93\% of slides annotated as lesion-free by the MIL model are in reality lesion-free.

\subsection{Segmentation Model}
\label{s:seg_model}

A multiclass segmentation model was trained to identify areas of necrosis, mitosis and parenchyma (healthy tissue) in image tiles using a pixel-level annotated subset of TGGates data \cite{igarashi2015open}. We refer to the Supplementary Information (SI) for further details concerning hyperparameters, training details and further architecture specifications. For the purpose of this paper, we will focus solely on the necrotic area prediction, and we will ignore the mitotic figure class.

Keeping in mind that the dataset is highly unbalanced, we report the precision, recall and area under the receiver operating curve (AUC) alongside the dice score (or F1 score) in Table \ref{tab:segmentation_data}. 
During evaluation, the tiles were masked such that only annotated pixels were included. This means that, as opposed to the training strategy, unlabelled pixels were not assumed to be healthy tissue (parenchyma).

\begin{table}[t!]
    \centering
    \begin{tabular}{c | c | c c c c c c c}
        \hline
         Model   & Dice Score($\uparrow$) & Precision($\uparrow$) & Recall($\uparrow$) & ROC AUC($\uparrow$)\\
         \hline
         \multirow{1}{*}{Segmentation} &  0.90 & 0.89 & 0.71 & 0.95 \\
         \hline
         \multirow{ 1}{*}{MIL} & - & 0.93 & 0.85 & $0.85$\\ 
        \hline
    \end{tabular}
    \caption{Summary of evaluation metrics for the segmentation model and the MIL model. 
    Note that the various metrics are computed at the pixel level for the segmentation model, while
    at the slide level for the MIL model. For the MIL model, we assigned the positive label to 
    lesion-free slides, in order to take into account the unbalanced nature of the dataset as well as the
    practical application of the model. }
    \label{tab:segmodel_results}
\end{table}

\subsection{Contextual explainability framework}

\begin{figure}
    \centering
    \includegraphics[width=0.9\textwidth]{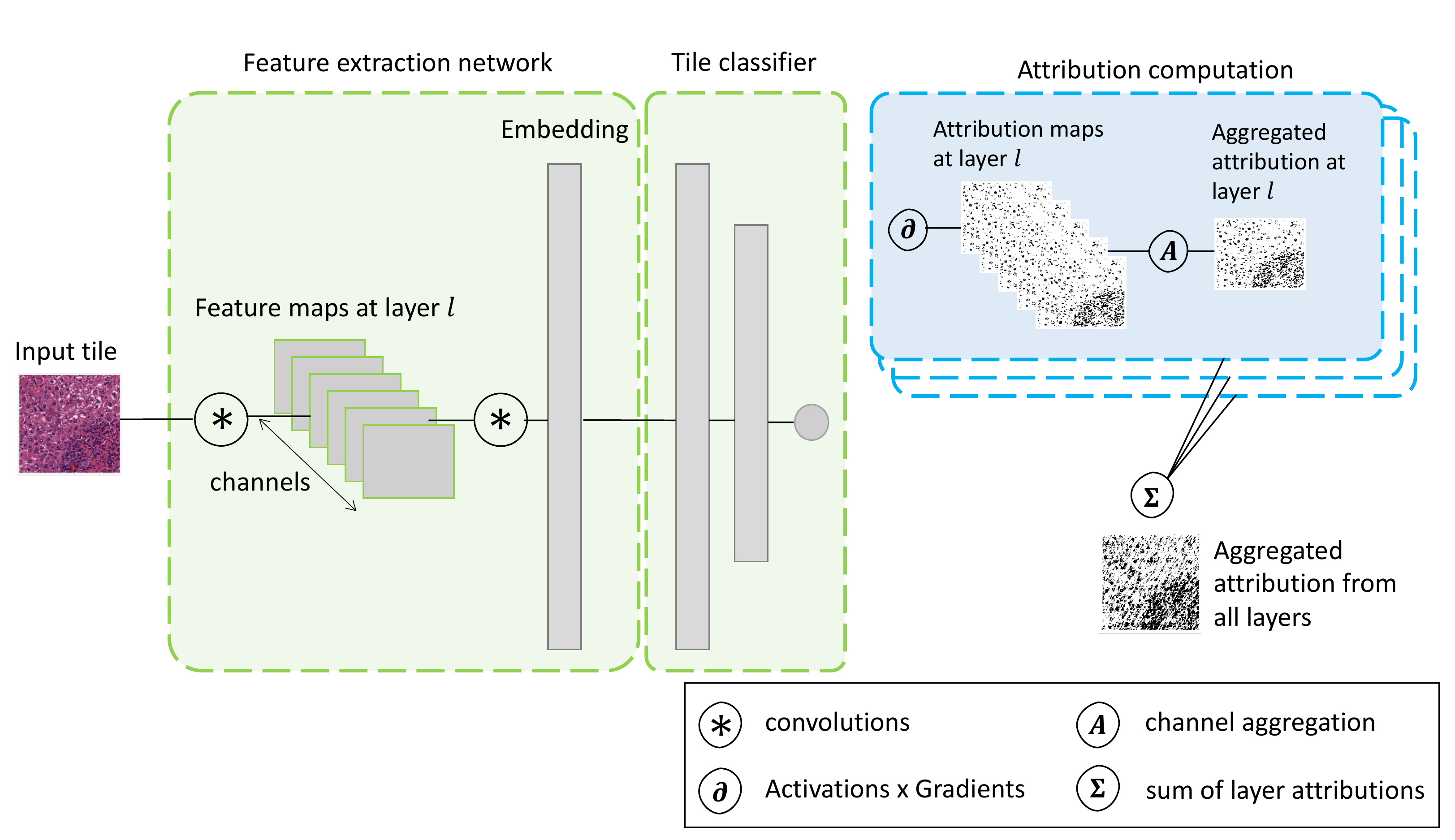}
    \caption{Overview of the explainability workflow. Feature extraction network and tile classifier network are fused, gradients are computed with respect to each channel for each convolutional layer of interest in the embedder network. The attributions maps are computed following the \emph{Activations} x \emph{Gradients} principle, multiplying the computed gradients by the respective activation map for each channel. Per convolution layer, channel attributions are aggregated (by taking the mean of absolute values or the variance), then all obtained activation maps are scaled back to the original input size and summed.}
    \label{fig:methods_schema}
\end{figure}

The classifier network $\Phi$ encodes, through its pretrained convolutional backbone, the morphological features of 
the input image into a lower-dimensional representation. It is well known that convolutional filters in early layers 
are sensitive to basic geometric features, e.g., angle and edge detectors; in contrast, more complex morphological features are responsible for filter activations in deeper layers \cite{olah2017feature}. 
We adopt the approach of \textit{contextual explanation} of \cite{bertolini_zhao_clevert_montanari_2022}, 
and use these layer attributions as a morphologically meaningful feature basis for our explanations. Figure \ref{fig:methods_schema} summarizes the different steps.

We represent the network $\Phi$ (our pre-trained feature extractor with the trained downstream network) as a composition of maps,
$\Phi : \cM \xrightarrow{\psi_l} \cM_l \xrightarrow{\xi_l} \R$, 
where $\cM_l$ is the output space of the $l{}^{\text{th}}$ convolutional layer. $\cM_l$ has dimension $k_l \times k_l \times C_l$,
where $C_l$ is the number of channels in the $l{}^{\text{th}}$ layer, and $k_l$ is the embedding size length in terms of superpixels.

Given a set of convolutional layers $\cCb = \{\cC_l\}$, where $l$ indicates the layer number, 
we compute layer-wise feature attribution values for $\Phi$,
choosing activation$\times$gradients to measure feature
importance. For a given layer $l$, we compute the attribution
\begin{align}
\label{eq:attrxgrads_lay}
    a_{l,c_l}(\bx) = \frac{\partial (\xi_{l,c_l})(\bx)}{\partial\psi_{l,c_l}(\bx)} \times \psi_{l,c_l}(\bx)~,
\end{align}
where the map $\psi_{l,c_l}$ is the restriction of the activation map $\psi_l$ to the $c_l{}^{\text{th}}$ channel in the layer $l$, 
and similarly for $\xi_{l,c_l}$.
As we do not wish to have one explanation for each channel of each layer, we propose to aggregate
the attributions \eqref{eq:attrxgrads_lay}.
In \cite{XAI4embeddings}, we showed how different aggregation strategies carry different properties. 
In particular, we showed how the mean absolute value (MAV) aggregation $\cA_{\text{MAV}} \equiv \mathbb{E}[|\bullet|]$
avoid the common problem of signal cancellation of the mean aggregation $\cA_{\text{mean}} \equiv \mathbb{E[\bullet]}$. 
Moreover, we also showed that the variance aggregator $\cA_{\text{var}} \equiv \Var[\bullet]$ 
leads to more dense attribution maps. 

For a given aggregation strategy $\cA$, we aggregate the channel-wise attributions \eqref{eq:attrxgrads_lay}
as follows:
\begin{align}
    a_l(\bx) = \cA[a_{l,c_l}(\bx)].
\end{align}
This aggregation produces a saliency map for each layer of the neural network. 
This allows us to extract a more complete picture of what the whole neural 
network has learnt during training. 
In fact, it has been shown that, while shallow layers focus on semantically simple and
spatially local features, deeper layers learn more complex and global features \cite{olah2017feature}. 
We aggregate attribution maps from different layers to then obtain a 
unique and contextual explanation of the model prediction, involving
both local and global morphological features. Specifically, we
combine the different layers attributions by
\begin{align}
    a(\bx) = \sum_{l\in\cC} a_l(\bx)~.
\end{align}
In practice, we will not compute attributions for each convolutional layer $l$ of $\Phi$. While doable for small models, it is certainly unfeasible for large models like ResNet50 in our case. Thus, we choose $\cC$ to be a subset of convolutional layers, which we will consider to be representative of the whole network. In order to access the different levels of learned concepts, we make sure to select layers uniformly along the network's depth. Our choice is to pick every 5-th convolutional layer, 
specifically $\cC=\{5, 10, 15, \dots, 50\}$. Thus, our explanations will contain local features, encoded in shallow layers, as well as global explanations, captured instead in deeper layers.

\begin{figure}[t!]
\center
\includegraphics[width=0.5\textwidth]{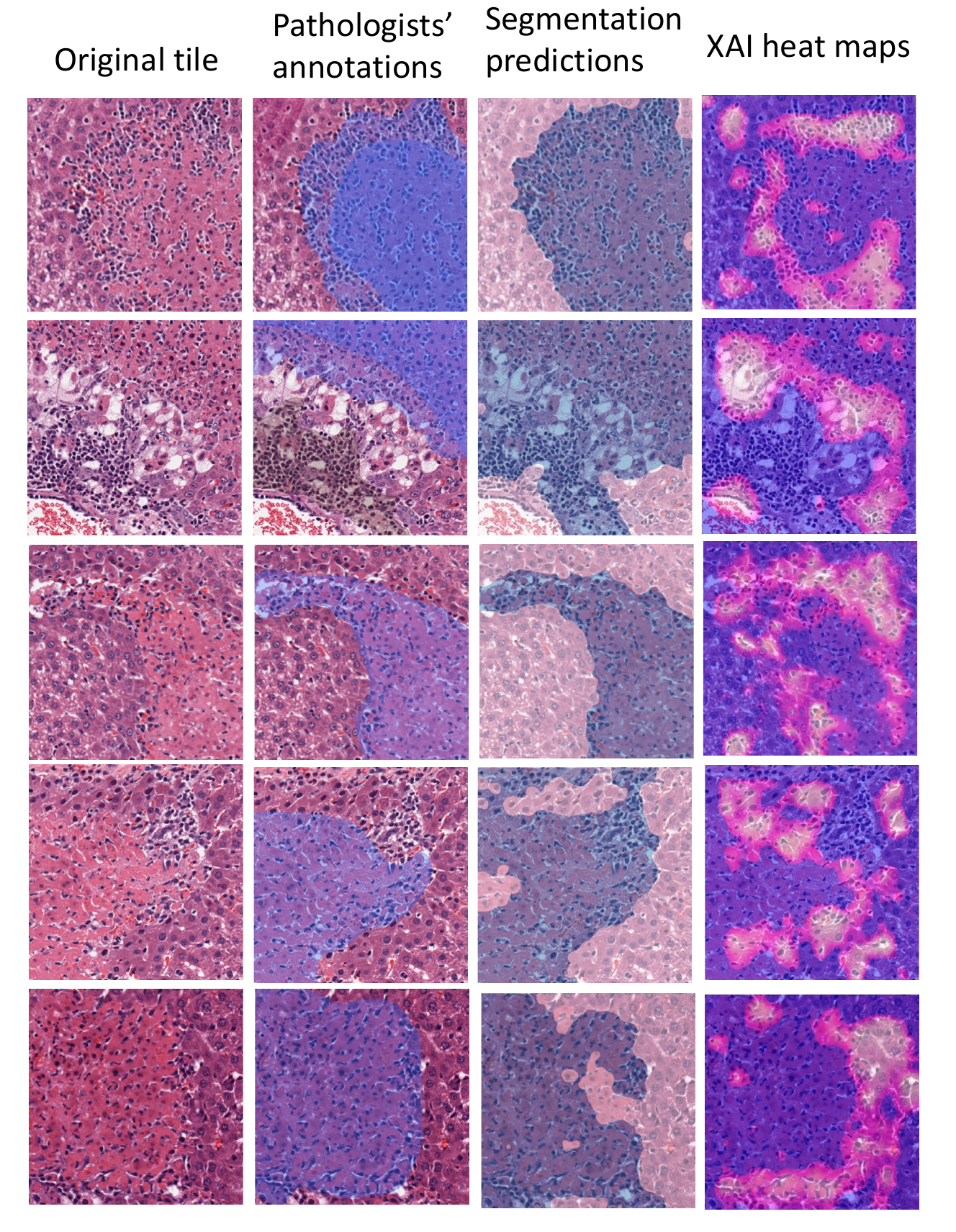}
\caption{Qualitative comparison of explanations with ground truth annotations and segmentation model predictions.}
\label{fig:quali}
\end{figure}

\subsection{Tile-level explanation normalization}
\label{ss:normalization}

Pathologists' annotations, which we will consider as ground truth when validating our
contextual explanations, are binary masks, assigning value $1$ to a region corresponding to the target
lesion and $0$ otherwise. 
Our explanations, instead, take value in the real numbers, i.e., $a(\bx)\in \R^{L \times L}$.
To deal with this issue, we normalize our attributions to be in the range $[0,1]$. Specifically,
we apply normalized percentile ranking to the explanation $a(\bx)$,
 defined as the percentage of how many values in $a$ are smaller than the current value,
that is,
\begin{align}
    (\at)_{ij}(\bx)=\frac{\sum_{mn} I(a_{ij}(\bx) > a_{mn}(\bx))}{L^2}~,
\end{align}
where $I$ is the Heaviside step function.
There is an additional rationale for normalizing attribution tile-wise, instead of, for
instance, slide-wise. Since our explanations are \textit{local}, i.e., 
they are computed with respect to a specific prediction of a given input,
it is unclear whether comparing attributions from different tiles is 
rigorously sound. 
If a given morphological feature in a tile was key for
the corresponding prediction, it does not imply it would be as important if
other, perhaps more relevant, features were also present in the image. 
We will see a manifestation of this phenomenon in Section \ref{s:experiments},
where we compare attributions obtained for different but overlapping tiles. There we observe that
a feature seen by both tiles will receive different importance, depending 
on what is present in the remaining of the images.

\section{Qualitative analysis of the explanations}
\label{s:qualitative}

We begin with some observations based on a qualitative assessment
of the explanations we computed. Such considerations are important since, 
ultimately, the primary goal of an explanatory framework is to 
equip the end user with the capacity to interpret the model's predictions. 

We often observe that, as illustrated in Figure \ref{fig:quali},
the model tends to focus on the border between the healthy tissue and the necrotic area. 
This means that the model has learned to detect
morphological and structural tissue changes (color, texture, cell structures, etc.) as a sign of lesion. 
That is, instead of learning the morphological properties characteristics of necrotic areas, the model rather learned
that such structural changes correlate with the diagnosis necrosis.
Perhaps surprisingly, this mirrors how human pathologists approach the reading of a slide. 
They take a first look at the slide at lower magnitude to obtain an overview of the tissue, and successively zoom in into areas which exhibit unusual  morphological characteristics.
Similarly, our explanations indicate that the MIL model did not necessarily learn how a necrosis area looks like, but that such changes in tissue occur only in positive slides. 
This is to be contrasted with the segmentation model's strategy, where the model is forced to detect all morphological features
defining the lesion in order to correctly solve the task.
A bit more abstractly, we can state that the space of possible learning 
strategies is inversely correlated with the amount of supervision we provide
to the model. A model with high level of supervision (as the segmentation model)
has only a limited choice in the way it can solve its task, due to the fine-grained
nature of the output required to be predicted (i.e., each pixel needs to be
correctly classified). Models with less supervision, instead, are more free to 
choose their route to correctly solve the task at hand, since their output is not forced
to be as detailed. For these models, explainability is especially important, since the model's
strategy will likely deviate from the humans' one, as, partially, is the case here. 

\begin{figure}[t]
\center
\includegraphics[width=0.95\textwidth]{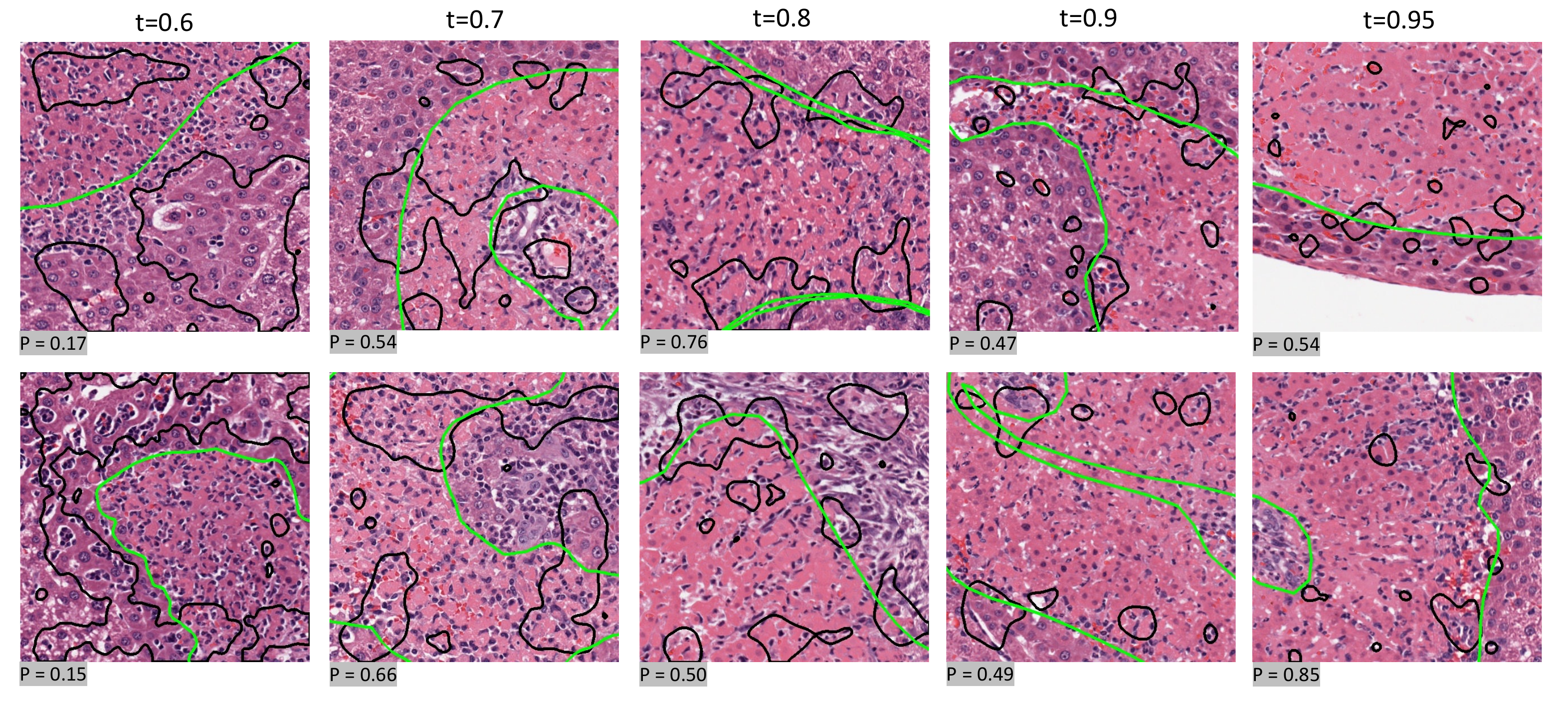}
\caption{Examples of explanations for different threshold values $t$ (in black). Pathologists' generated annotations are depicted in green. Best seen in color.}
\label{fig:gt_ex}
\end{figure}

This qualitative analysis of our explanations raises the issue that explanations themselves are not trivial to interpret. 
Their meaning is, in fact, strictly dependent on the algorithm of the model to be explained. 
If the attribution maps were, for instance, the output of a segmentation model, we would instantly conclude
that the model is not performing the task correctly. 
Thus, in order for the explanations to be a useful tool in the hand of the pathologist, 
it is crucial that they are aware of the model's learning objective.
This is key in order to avoid looking at explanations
just as a way to confirm the experts' knowledge, but rather as a meaningful
tool to understand how the model approaches the predictive task. 

In any case, if the model is correctly trained, the explanations should be meaningful and
should correlate, in some way, to the pixel-level ground truth. In the next section, we
explore this correlation for our contextual explanations in a more quantitative way.

\section{Experiments}
\label{s:experiments}
In this section, we turn to a more quantitative evaluation of our 
XAI framework. We focus here on two important aspects. First,
we evaluate the accuracy of our explanations by quantifying the agreement with
manual annotations from pathologists (obtained for a subset of test slides), as well as with the predictions
of the segmentation model trained on those annotations. 
Second, we evaluate the stability of our explanations by proposing a measure of
robustness. Namely, we compute attribution maps for overlapping but not coincidental tiles, and we compute the agreement of the attribution maps on the overlapping area.

\subsection{Set-up and scores}

\begin{figure}[t]
\includegraphics[width=\textwidth]{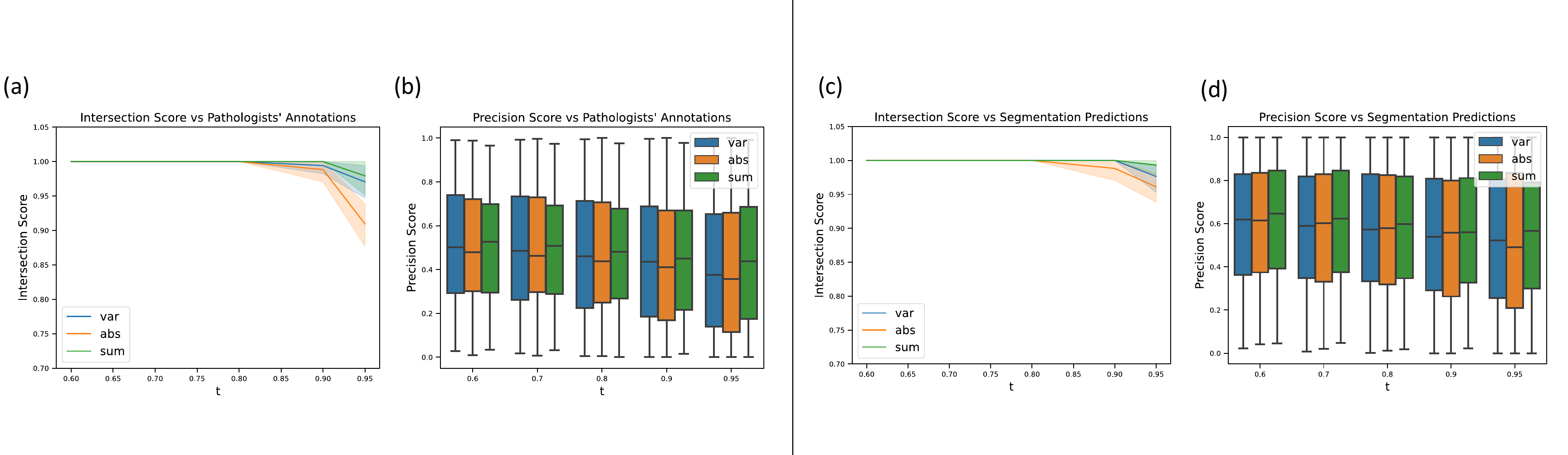}
\caption{Quantitative comparison of explanations with ground truth annotations from pathologists (a,b)
and segmentation prediction annotations (c,d) for different explanation thresholds $t$ and channel aggregation strategies $\cA$.}
\label{fig:groundtruth}
\end{figure}

Since the ground truth annotations assume a binary value ${0,1}$ ($0$ if the pixel
is not annotated, $1$ if it is), it is challenging to calculate the agreement with 
the float-value attribution maps. Thus, we choose to compute a
binary mask obtained by thresholding the (already normalized
as described in section \ref{ss:normalization}) attribution maps. 
Given a certain threshold value $t\in [0,1]$, we have
\begin{align}
\label{eq:thresholding}
    a_t(\bx) = \begin{cases}
    0 & \text{if } a(\bx) < t~,\\
    1 & \text{if } a(\bx) \geq t
    \end{cases}~.
\end{align}
Thus, $a_t(\bx)$ depicts a mask selecting the top $100\times(1-t)\%$ pixels
according to the values of the attribution map. 
Obviously, the choice of a value for the threshold $t$ is arbitrary, thus we will
perform many of the experiments below for various values of $t$. We will show that
our results are robust with respect to the choice of $t$ and that, for instance,
we do not achieve a good overlapping with the ground truth annotations
simply by selecting a low threshold (thus activating ``enough" pixels).
Some examples of thresholded explanations for several values of $t$ are
reported in figure \ref{fig:gt_ex}.

We compute a number of statistics which are designed to measure the
accuracy of our explanations. The most basic question is whether 
the explanations $a_t$ intersect at all the ground truth annotations $g(\bx)$. 
The corresponding score takes the form
\begin{align}
\label{eq:intersection_score}
    I(a_t, g) = \mathbb{E}_{\bx\sim\cD_{\text{test}}}[\max(a_t(\bx)\odot g(\bx))]~,
\end{align}
where $\odot$ corresponds to element(pixel)-wise multiplication
and the expectation value is computed over all tiles in the test dataset. 
In the case that $g$ represents pathologists' annotations, since the slides are not fully annotated, we exclude test tiles in which
no target annotation is present. 
When comparing our explanations to the segmentation model outputs, $\cD_{\text{test}}$
spans all the tiles of the test slides, since we can simply run inference on all of them.
Note that $a_t(\bx)\odot g(\bx)$ is a $512\times512$
matrix with values $1$ for pixels in the intersection of the annotations, and $0$ otherwise. 

In order to get a better quantification of the overlap between the 
explanations $a_t$ and the ground truth annotations $g(\bx)$, we also compute the area of intersection between them. We propose a score consisting of the (normalized) intersection area $\text{sum}(a_t(\bx)\odot g(\bx))$.
The score
\begin{align}
\label{eq:precision_score}
    P_t(a_t, g) = \frac{1}{(1-t) L^2}\mathbb{E}_{\bx\sim\cD_{\text{test}}}\left[\text{sum}(a_t(\bx)\odot g(\bx))\right]~,
\end{align}
where $L=512$ is the tile size, computes the average (over the test dataset) percentage of explanation area that overlaps with the ground truth annotations. 
That is, we can interpret this as the precision score (True Positives / Predicted Positives) for our attribution map. 
In figure \ref{fig:gt_ex} we reported the precision scores below each example tile.

Another useful score which is often used in the literature when evaluating agreement of annotations is the Intersection over Union (IoU) score
\begin{align}
\label{eq:iou}
    \text{IoU}_t(a_1(\bx), a_2(\bx)) = \mathbb{E}_{\bx\sim\cD_{a_1\cap a_2}}
    \left[\frac{\text{sum}(a_1(\bx)\odot a_2(\bx))}{\text{sum}(a_1(\bx) + a_2(\bx) - a_1(\bx)\odot a_2(\bx))}\right]~.
\end{align}
We will make use of this score in section \ref{ss:stability} when studying the agreement of 
explanations on different but overlapping tiles. This score is not suitable to compare our
explanation with pixel-level annotations because we are thresholding our 
attribution maps as in \eqref{eq:thresholding}.

\subsection{Agreement with pixel-level annotations}
\begin{figure}[t]
\center
\includegraphics[width=\textwidth]{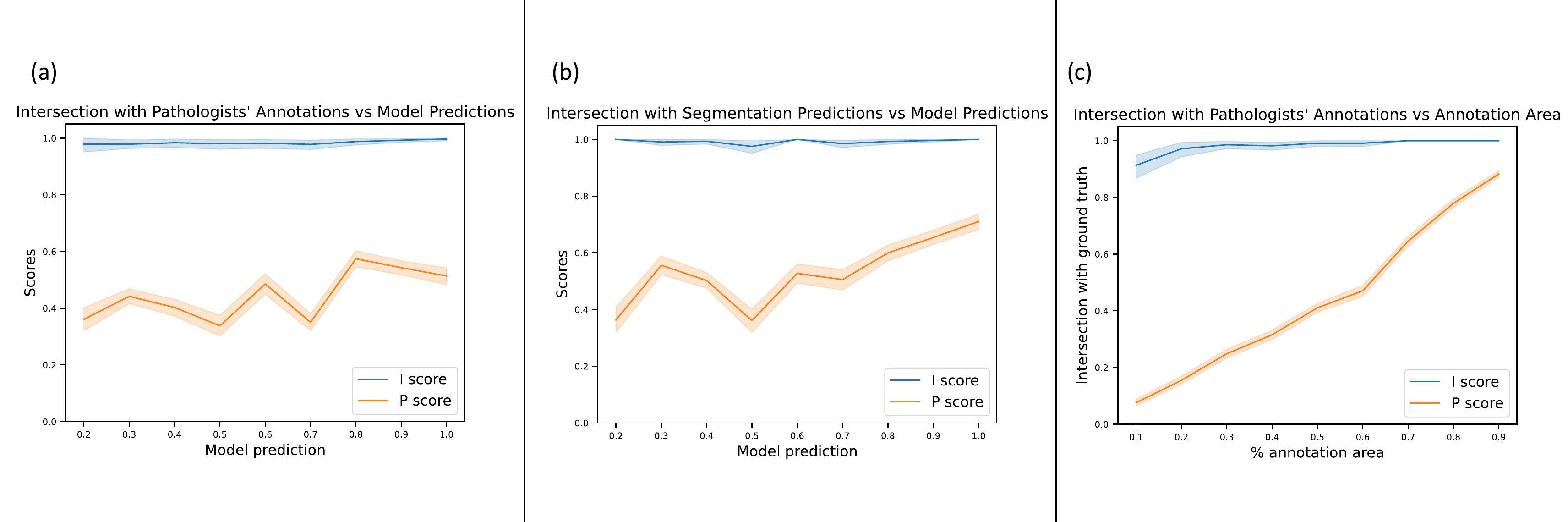}
\caption{Scores as a function of model prediction (a,b) and of annotated area (c).}
\label{fig:acc_vs_pred}
\end{figure}

In Figure \ref{fig:groundtruth} we depict the scores \eqref{eq:intersection_score}
and \eqref{eq:precision_score} defined in the previous section for various values of the threshold $t$, where $g$ represents the ground truth annotations from pathologists (a-b) or from inference from the segmentation model (c-d). 
We report these scores for three attribution channel aggregation strategies, namely $\texttt{abs}$, $\texttt{mean}$ and $\texttt{var}$. 
We observe an almost identical performance for all aggregation strategies. We interpret this as a sign of robustness of our explanation framework and
a confirmation of the signal from the model prediction. 

For all scores, we observe almost no decrease in value when increasing the threshold, thus decreasing the number of ``explanation pixels'', according to \eqref{eq:thresholding}. 
In particular, Figure  \ref{fig:groundtruth}a shows that even for a threshold $t=0.95$, that is, only $5\%$ of pixels are selected for our explanations, $a_t$ intersect $g$ in more than $90\%$ of the cases.
This provide strong evidence that we do not just achieve an overlap with the ground truth by ``highlighting enough pixels'' in our explanations.

Figure \ref{fig:groundtruth}b depicts the precision score \eqref{eq:precision_score} as a function of the threshold. Also here we observe only a minimal decrease in the value of the score for increasing threshold $t$. Slightly more evident is rather an increase in the standard deviation, rather than a decrease of the mean value for the score. 
For all threshold values and aggregation strategies, the mean value for the score is between $0.4$ and $0.5$, which means that, in average, $40-50\%$ of the attribution map overlaps
with the pathologists' annotations. We interpret this as a quantitative
confirmation for our qualitative discussion in section \ref{s:qualitative}, where we
pointed out that the model seem to solve the learning task by recognizing the boundary
region between the healthy and necrotic tissue. This is indeed consistent with
a prediction score of around $0.5$: in fact, by highlighting the border tissue, the explanations will be roughly split evenly between the two ground truth annotation classes.

In Figure \ref{fig:groundtruth}c-d we reported the scores using as ground truth $g$ the prediction of
the segmentation model trained on the pathologists' annotations as described in Section \ref{s:seg_model}.
The conclusions from the previous section apply here as well: all scores are mostly unaffected by
the chosen value of the threshold, and even at $t=0.95$ we observe an average intersection rate of
$I>0.95$ and precision score of $P\sim 0.55$.

In figure \ref{fig:acc_vs_pred}a-b we report the scores above as a function of the model prediction for the tile. Here we include explanations for all thresholds. We observe that
all scores correlate positively with the model prediction. This property is referred to as \emph{faithfulness} in \cite{sanchez-lengeling2020} and is one of the metrics the authors propose to evaluate the quality of an explanation.  Here, as expected, the correctness of the explanations increases with the certainty of the model about the prediction. 
This property is a desired feature, as it validates the assumption that the model making
a very certain prediction (output value close to $1$) accurately identifies the correct morphological features. On the other hand, when the model is less certain about a prediction, 
it seem to be less accurate about recognizing the key features in the tile image. 
At the same time, this is an additional confirmation of our explanatory methods, which 
is able to precisely identify the key components, when the model is very sure that the tile contains a necrosis lesion. 
 
Since the ground truth area varies from tile to tile and the explanation area is instead fixed due to thresholding, we expect a positive correlation between the scores and the annotated area, which is what we observe in figure \ref{fig:acc_vs_pred}c. The intersection score \eqref{eq:intersection_score} assumes an average value of 
over $0.9$ even when only $10\%$ of the tile is assigned to a necrotic area.

\subsection{Stability of the explanations}
\label{ss:stability}

\begin{figure}[t]
\center
\includegraphics[width=0.95\textwidth]{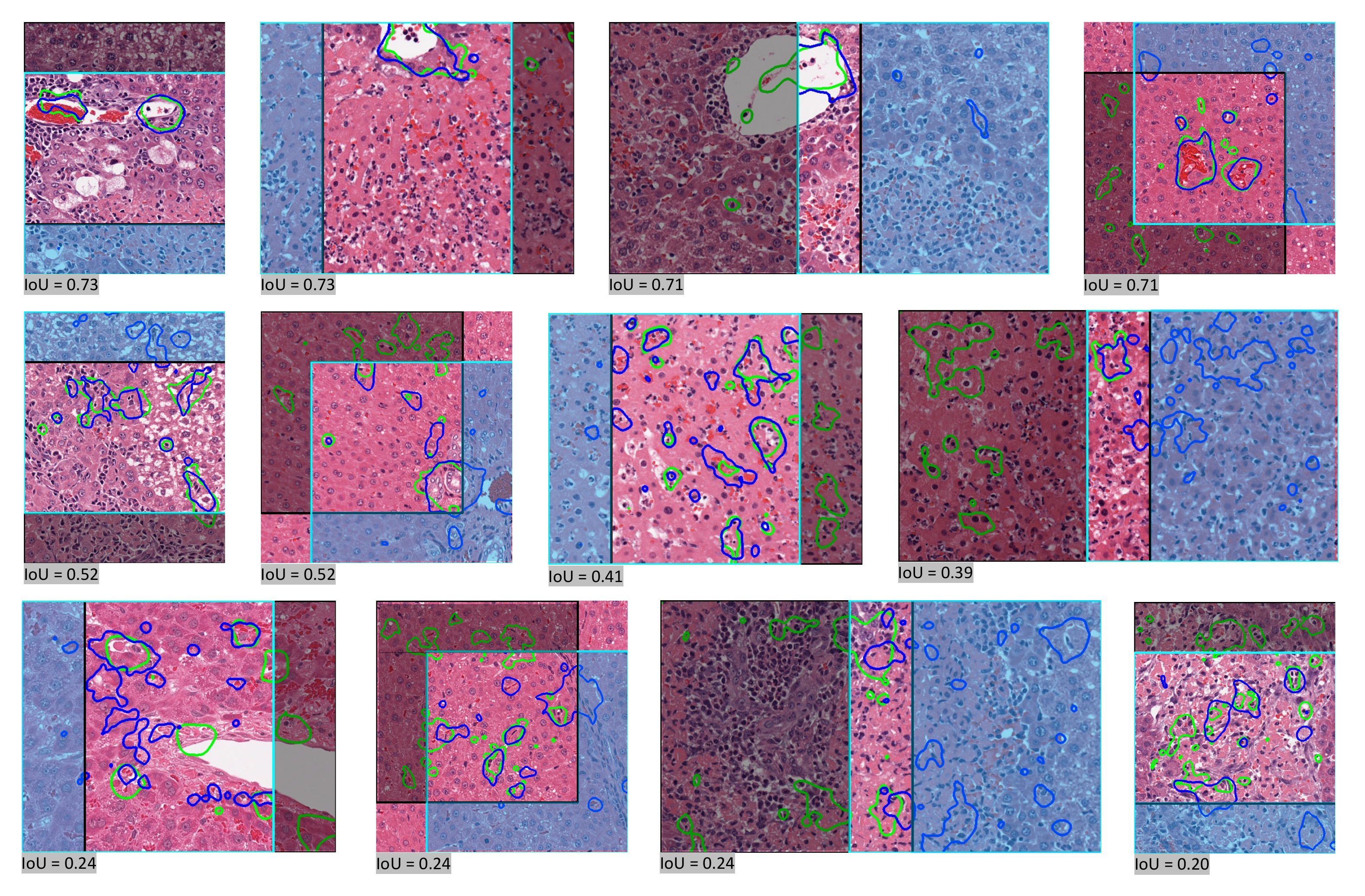}
\caption{Examples of explanation for overlapping tiles. The black and turquoise squares corresponds to the 
overlapping tiles. The green and blue annotations represented the corresponding thresholded explanations 
($t=0.9$). Best seen in color. }
\label{fig:overlap_ex}
\end{figure}

Next, we turn to a study of robustness of our explanation. Namely, we generate
predictions and the corresponding explanation attributions for overlapping tiles, and we
quantify the degree of correlation between the two explanations. 
Specifically, for each slide in the test set, we generate 3 additional grids, which are shifted by
$\mathbf{\delta} = \delta_x L/4 , \delta_y L/4$, where $\delta_x, \delta_y \in \{0,1\}$. We then construct a dataset
consisting of all the pairs of tiles from different grids which exhibits a non-zero intersection area. 
Note that these combinations of shifted locations result in different overlapping areas between a tile and its shifted neighbours.
Let us denote $a_t$ and $a_t^{\bdelta}$ the corresponding attributions. 
We compute 
the IoU score \eqref{eq:iou} to quantify the degree of agreements of the explanation on the 
overlapping area, that is, we ignore the explanation outside of it. 
Note, however, that the attributions are normalized and thresholded with respect
to their original area of prediction. This means that we do not expect a perfect agreement, 
as it is conceivable that an important morphological feature is located outside the
overlapping area and correctly captured by one attribution, which is therefore
de-prioritizing features commonly observable by both maps. 

Figure \ref{fig:overlap_ex} shows some qualitative examples of such overlapped attributions,
for the threshold $t=0.9$. The two distinct 
original tiles are depicted with black and turquoise boundaries, respectively. 
We report the IoU scores with each example to provide some intuition about the values. 
In the top row, we observe that IoU values of about $0.7$ correspond to almost perfect 
agreement between the attribution maps. The middle row shows some examples of about the average value
of $\sim0.4-0.5$. Also in this case we observe a good agreement, as almost all areas in the 
overlap region are highlighted by both maps. In comparison with the top row, we see that the boundaries of the regions do not coincide to a high level of precision. 
Finally, we show in the last row some examples with rather low scores of about $\sim 0.2$. 
Even in this case, we see some decent correlation between the two sets of 
XAI-generated annotations. However, in this case we see that some areas are
missed by one of the two attributions.

\begin{figure}[t]
\center
\includegraphics[width=1.0\textwidth]{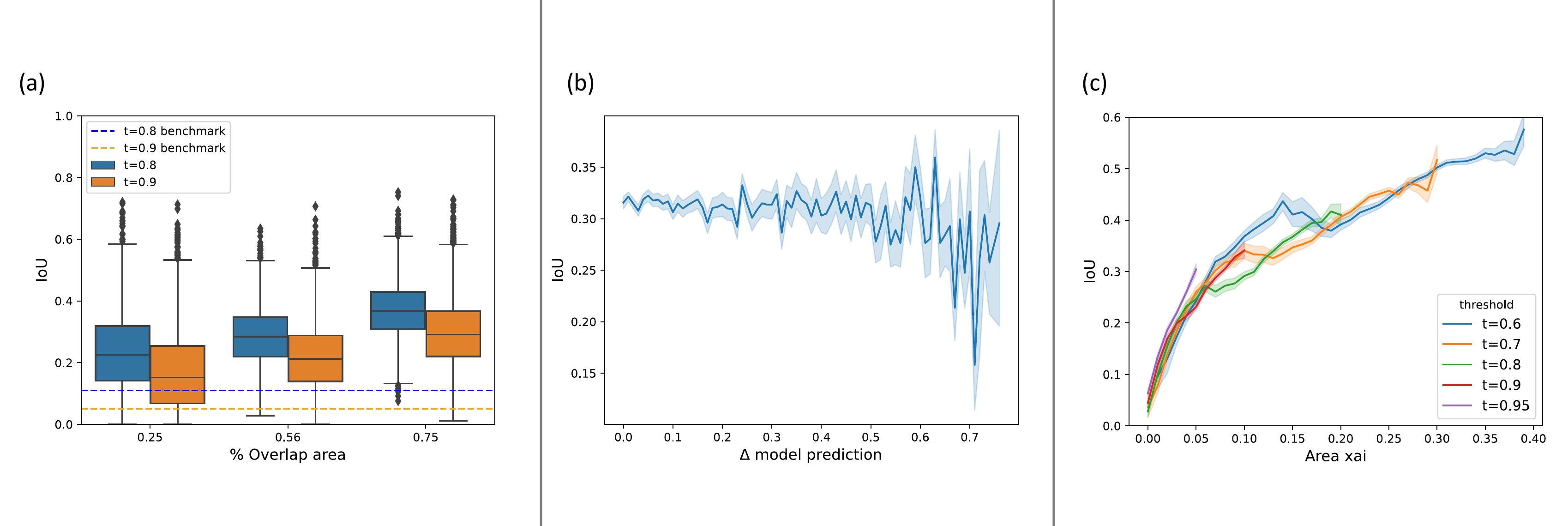}
\caption{IoU scores as a function of the overlapping area (a), of the difference of model prediction between the two overlapping tiles (b), and of the xai annotation area in the overlapping region (c).}
\label{fig:overlap_scores}
\end{figure}

In Figure \ref{fig:overlap_scores}a we report the IoU score \eqref{eq:iou}
for two thresholds $t=0.8, 0.9$ and
for different tile overlapping areas. We observe, as expected, that the score correlates positively
with the area of the overlapping region. This is expected since, as mentioned above, 
the smaller the overlapping area, the higher the probability that the attribution maps
find and highlight relevant morphological features outside of it. 
We also depicted as a dotted line a benchmark represented by random heatmaps 
following a uniform distribution $\cU(0,1)$.
We derive these benchmark values for any threshold $t$ in the SI.

In Figure \ref{fig:overlap_scores}b we depict the IoU score as a function of the difference in 
model predictions for the two overlapping tiles. We observe that the IoU score is fairly independent of
the model prediction, for differences in prediction score of up to 0.5 (after which the amount of cases with larger shifts in prediction scores are too low to draw conclusions). We note however, that we restricted ourselves to tiles with initial predictions above $0.2$,
thus for which the model assigned some signal for the positive class. Thus, we can conclude that
the the model explanations are robust even in cases when the model certainty about the prediction 
considerably drops. 

Finally, Figure \ref{fig:overlap_scores}c shows the IoU score as a function of the area of the 
$a_t$ annotations in the overlapping region, for different thresholds. We observe, as expected, 
that the scores increases with the amount of annotated area in the overlapping region. 
This facts confirms what we argued for above: the disagreement between the explanations
is more prominent when one map focuses its attention to features which are
not seen by the other due to the shift.

\section{Conceptual integration within an industry workflow}

\begin{figure}[t]
\includegraphics[width=\textwidth]{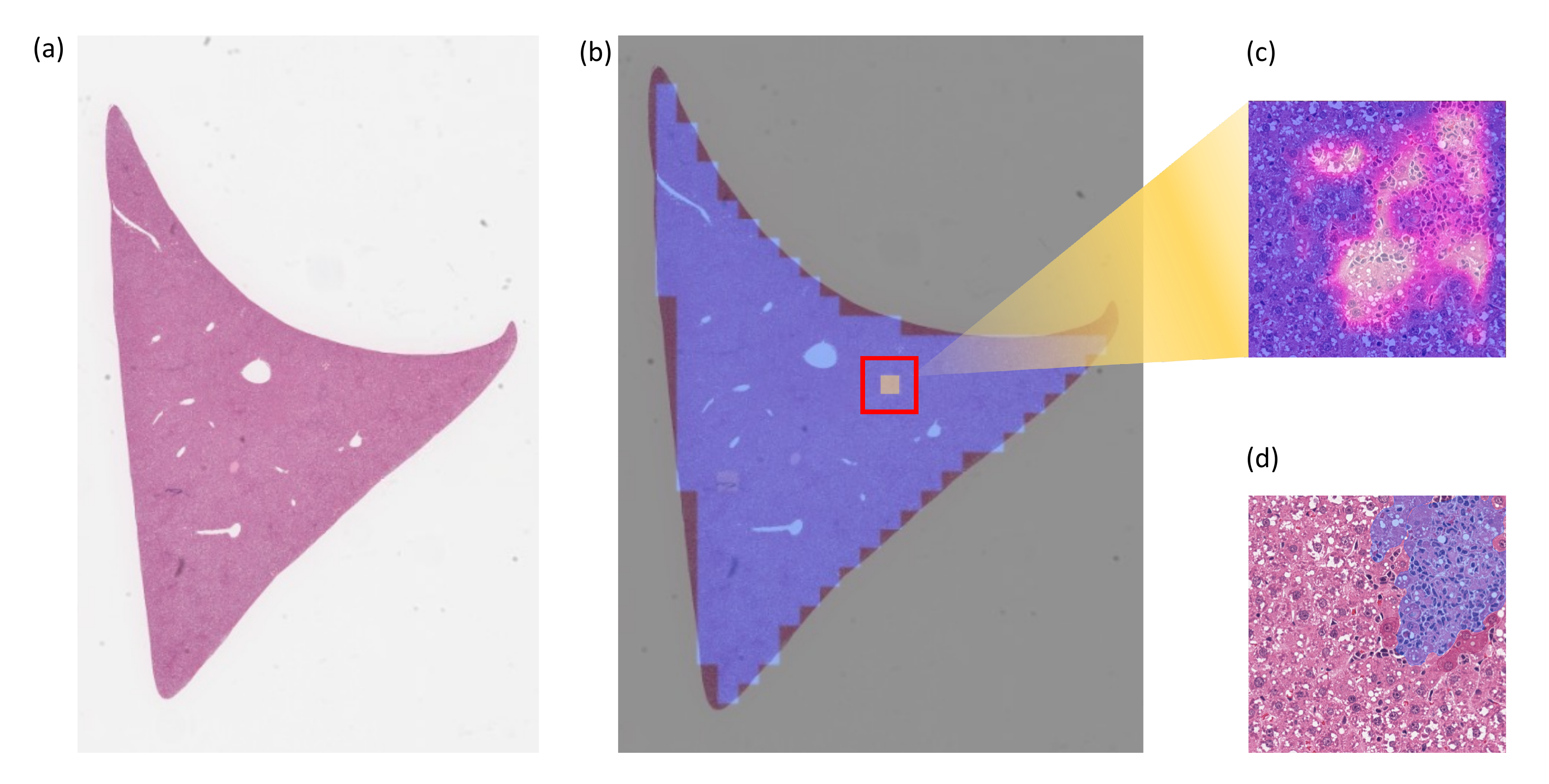}
\caption{From slides through tiles to pixels: an example of weakly-supervised models' evaluation with contextual explanations.
(a) represents a new slide which a pathologist needs to read in the given study. (b) is the visualization of the output of the 
MIL model, in which each tile is scored (blue represents a low score, yellow the prediction of presence of the lesion). The
pathologist can then focus on the top tiles scores by the model and access the contextual explanations (c). We reported in 
(d) the ground truth annotation for comparison purposes (these are obviously not available to pathologists when reading a new study).}
\label{fig:pado_vis}
\end{figure}

We have presented tools that can help make sense of weakly-supervised models' predictions on individual WSIs. We would like now to place this work back into the context of a pre-clinical study analysis, as performed in the pharmaceutical industry. The workflow is presented in figure \ref{fig:pado_vis} and it is implemented at Bayer with the \texttt{Pado Visualize} package \cite{pado_vis}. 
The first step is the digitization of the physical microscopy slides prepared according to GLP standards, to acquire a study-level dataset of WSIs. This dataset contains metadata such as the organ, animal, dosage information related to each image. We would then propose to send all slides corresponding to rat livers to the necrosis MIL model, which would label them with a score reflecting the likelihood for each slide to contain a necrosis lesion. Using this score to rank the images, the pathologist in charge of the study could already detect enrichment of high-scored tiles in higher dosage regimen, for example, and make assumptions about toxic doses for the tested compound. By focusing first on the high-scored slides, they can optimize their workload. Once loaded into a visualization software, a slide that has been highly ranked by the MIL model will contain highlighted patches: these particular patches correspond to individual tiles which had the highest scores according to the model. How many such tiles to visualize can be decided by the user. This further allows them to recognize regions of particular interest, where a deeper look is needed. For specific tiles of interest, they can then enable the contextual explainability workflow we presented here, letting the model highlight in high-resolution the areas in the tile that led to such a score by the model. We already saw that these highlights do not correspond exactly to how a pathologist would manually delineate healthy from necrotic tissue, but can however shed some light on the model's decision process. We leave the presentation of such an integrated analysis and visualization workflow to future work.

\section{Conclusion}

Data in Digital Pathology is both a blessing and a curse. On the one hand, 
there is a growing effort of scanning and digitalizing histopathological slides, resulting in
vast datasets. In addition to the TG-GATEs dataset, we leveraged for our work the BigPicture Project, a public-private partnership funded by the EU Innovative Medicines Initiative (IMI), which for instance, aims to collect a repository of about 3 million WSIs for the development of AI algorithms. On the other hand, providing consistent and standardized metadata is a challenge, due
to the very technical nature of the field. Obtaining pixel-level annotations is even more costly, as such data
needs to be generated specifically for an AI application, and is not part of the standard pathologists' reports. 

In this paper, we presented a collection of approaches for a comprehensive ML pipeline for pre-clinical
Digital Pathology, leveraging the amount of available image data
and the scarcity of pixel-level annotations.
We adapted a recently proposed XAI technique to a MIL model trained to detect necrosis in 
rats' liver.
The explanations our methods generate enhance weakly supervised models' predictions
with pixel-level heatmaps. We compared these to pathologists' annotations and
segmentation models' predictions, and showed that our explanations positively correlate with
both ground truths. 
However, the learning task of the MIL model differs fundamentally from the one of segmentation models. The latter is trained to detect pixels that pathologist annotated as findings. The former
needs to agree with the pathologists' diagnosis only at the slide level. 
Our explanations gave us insights into how the model chose to solve the task. The model
recognized that the presence of morphological shifts correlates with the diagnosis label. 
These shifts in color, texture and cell morphology correspond to the surface boundary between healthy and necrotic tissue.

This work represents a first step in the direction of Digital Pathology at Bayer, where we will put into the hands of certified pathologists some tools to prioritize their work. The approaches are meant as enhancement for human diagnosis, and never as replacement. Models for many more lesions and organs are being developed in-house to increase the reach and usefulness of our ML toolbox.

\section{Acknowledgements}
The authors acknowledge funding from the Bayer AG Life Science Collaborations “Explainable AI” and “PathDrive”. This project has received funding from the Innovative Medicines Initiative 2 Joint Undertaking under grant agreement No 945358. This Joint Undertaking receives support from the European Union’s Horizon 2020 research and innovation program and EFPIA (\url{www.imi.europe.eu}, \url{www.bigpicture.eu}). The authors additionally thank Hannah Pischon and Moritz Radbruch, pathologists at Bayer, for the helpful discussions and feedback on model predictions and explanations.

\bibliography{references.bib}
\newpage
\appendix

\section{MIL Model Methods}

The model consists of a composition of two networks. We use a pretrained ResNet50 \cite{he2016deep} 
for feature extraction, where we remove the classification layer. Recall that
the weights of this network are not updated during training. The classification network,
which takes as its input the output of the ResNet50, consists of a stack of 2 fully connected layers of size 2048 and 1024 respectively. We use ReLU has the non-linear
activation function. The output layer consists of a fully connected layer with two 
outputs, corresponding to the scores assigned to the two classes. To these we apply 
a Softmax function and we train the model with the Cross-Entropy Loss. 

\section{Segmentation Model Methods} 
\label{sec:segmentation_methods}

A standard U-Net \cite{unet} was used to perform semantic segmentation of the tiles. The Segmentation Models library \cite{Iakubovskii:2019} allowed us to use a ResNet \cite{he2016deep} backbone pretrained on ImageNet as the encoder. The encoder, or contracting path, transforms $512\times512\times3$ dimensional tiles into a representation of $16\times16\times256$ in five stages, generating features of reduced spatial dimensions and increasing depth at each stage. The decoder, or expansive path, then transforms the encoder's final representation into a $512\times512\times C$ dimensional representation (a mask with $C$ channels, where $C$ is the number of classes). It does this by using ``up-convolutions" to gradually reduce the number of feature channels, while simultaneously increasing the spatial dimensions. At each stage in the decoder, the representations extracted by the encoder with matching spatial dimensions are incorporated. This information sharing between the contraction and expansion paths through what are known as ``skip connections" are crucial to ensuring that the final $C$-channel output contains accurate spatial information. Finally, a Softmax non-linearity is applied along the channel axis of the output compressing the raw logits into the range $[0,1]$. At each of the $512\times512$ pixel positions in the  output representation, the channel vector $\in \mathbb R^C$ represents the probability distribution of that pixel belonging to one of the $C$ classes.

The network was trained for 50 epochs using a ``dice-loss" function. This loss function, introduced by Milletari et al \cite{milletari2016v} and based on the Sørenson-Dice coefficient, was chosen to reduce the effects of class imbalance. Histopathology images often contain small percentages of lesion compared to healthy (or, in our case, unlabelled) tissue. Attempting to directly maximize classification accuracy is likely to trap the model in a local minimum where the model is highly biased towards labelling every pixel as normal. To penalize mis-classified lesions as well as normal tissue, Milletari et al designed a loss function which is proportional to the overlap of the predictions with the ground truth for each class. This value ranges between 0 and 1, where 0 represents a perfect prediction (opposite to the actual Dice coefficient). For a single class, where the predicted probability for pixel $i$ is $p_i$ and the ground truth class value is $g_i$, the Dice Loss over $N$ pixel predictions is
\begin{align}
\label{eq:dice_loss}
    \text{D}_c = 1 - \frac{2 \sum_i^N p_i g_i}{\sum_i^N p_i^2+ \sum_i^N g_i^2}
\end{align}
In a multiclass setting, we avoided learning a class weighting function \cite{crum2006generalized} and simply took the average across the output channels 
$\frac{1}{C}\sum_{c\in C}D_c$.

\section{Overlap threshold for uniform distribution}

In this section we derive the average intersection scores 
for ``random'' attribution maps. Let $\br,\bs \sim \cU(0,1)^{L\times L}$, that is 
a $L\times L$ matrix, where each entry is a random variable following
a uniform distribution between the values 0 and 1. We denote $\br_t$ and $\bs_t$ the
corresponding thresholded quantities, as defined in equation \ref{eq:thresholding} from the main text.
Since the entries in the maps are independent from each other, we can restrict ourselves to compute the scores component-wise. The results will hold for all components. 
First, we have
\begin{align}
    \mathbb{E}_{r\sim\cU(0,1)} [r_t] = 1-t~,
\end{align}
which follows directly from the definition of a uniform distribution. Thus, for the intersection we have
\begin{align}
    \mathbb{E}_{r,s\sim\cU(0,1)} [r_t \times s_t] = (1-t)^2~,
\end{align}
since the two random variables are iid. For the union instead we have
\begin{align}
    \mathbb{E}_{r,s\sim\cU(0,1)} [r_t + s_t - r_t \times s_t] = 2(1-t) - (1-t)^2 = (1-t)(1+t)~.
\end{align}
Finally, putting all together we have the IoU score
\begin{align}
    \text{IoU}(r_t,s_t) =\frac{1-t}{1+t}~.
\end{align}
The precision score (equation (8) in the main text) instead takes the value
\begin{align}
    P_t(r_t, s_t) = 1-t~.
\end{align}

\end{document}